\documentclass[conference]{IEEEtran}
\IEEEoverridecommandlockouts

\usepackage{cite}
\usepackage{amsmath,amssymb,amsfonts}
\usepackage{algorithmicx}
\usepackage{algpseudocode}
\usepackage{float}  
\usepackage{algorithm}
\usepackage{multirow}

\usepackage{graphicx}
\usepackage{textcomp}
\usepackage{xcolor}

\usepackage{hyperref}

\usepackage{orcidlink}

\hypersetup{hidelinks, colorlinks=true,allcolors=black,pdfstartview=Fit,breaklinks=true}

\def\BibTeX{{\rm B\kern-.05em{\sc i\kern-.025em b}\kern-.08em
    T\kern-.1667em\lower.7ex\hbox{E}\kern-.125emX}}
    
\begin{document}

\title{CSCE: Boosting LLM Reasoning by Simultaneous Enhancing of Causal Significance and Consistency
\thanks{This work was supported in part by National Nature Science Foundation of China (No. 62172036) and National Science and Technology Major Project (2022ZD0116305). Accepted for presentation at IEEE International Conference on Multimedia \& Expo 2025.}
}
\author{
    \IEEEauthorblockN{
        Kangsheng Wang\orcidlink{0009-0009-8392-4148}, 
        Xiao Zhang\orcidlink{0009-0007-2804-6915},
        Juntao Lyu\orcidlink{0009-0006-3374-047X},
        Tianyu Hu$^{*}$\orcidlink{0000-0001-9903-0696} \thanks{$^{*}$ Corresponding author.},
        Huimin Ma\orcidlink{0000-0001-5383-5667}
    }
    \IEEEauthorblockA{
        \textit{School of Computer and Communication Engineering} \\
        \textit{University of Science and Technology Beijing}\\
        Beijing, China \\
        jackie@ieee.org; \{D202310433, lyujuntao\}@xs.ustb.edu.cn \\
        \{tianyu, mhmpub\}@ustb.edu.cn
    }
}

\maketitle

\begin{abstract}
Chain-based reasoning methods like chain of thought (CoT) play a rising role in solving reasoning tasks for large language models (LLMs). However, the causal hallucinations between \textit{a step of reasoning} and \textit{corresponding state transitions} are becoming a significant obstacle to advancing LLMs' reasoning capabilities, especially in long-range reasoning tasks. This paper proposes a non-chain-based reasoning framework for simultaneous consideration of causal significance and consistency, i.e., the Causal Significance and Consistency Enhancer (CSCE). We customize LLM's loss function utilizing treatment effect assessments to enhance its reasoning ability from two aspects: causal significance and consistency. This ensures that the model captures essential causal relationships and maintains robust and consistent performance across various scenarios. Additionally, we transform the reasoning process from the cascading multiple one-step reasoning commonly used in Chain-Based methods, like CoT, to a causal-enhanced method that outputs the entire reasoning process in one go, further improving the model's reasoning efficiency. Extensive experiments show that our method improves both the reasoning success rate and speed. These improvements further demonstrate that non-chain-based methods can also aid LLMs in completing reasoning tasks.
\end{abstract}


\section{Introduction}
Large language models (LLMs) have shown promising potential in solving structured logical reasoning problems\cite{logi1,logi2,logi3}, unstructured mathematical reasoning problems\cite{math1,math2,math3} and downstream tasks such as autonomous driving \cite{zhang2025enhancing,hu2025autonomous}. Previous research indicates that although chain-based methods, represented by Chain of Thought (CoT)\cite{CoT}, can effectively enhance LLMs' reasoning capabilities, they cannot avoid the inherent issue in the answer generation mechanism of LLMs—namely, the causal hallucination between a step of reasoning and corresponding state transitions, between which the causal relationship is not significant or consistent\cite{non-causal}. 

CoT reasoning enhances the coherence of LLM inferences but struggles to address the issue of causal hallucination, as it relies more on pattern recognition than on deep causal inference\cite{wei2022chain}. In an experiment we implemented, we estimated the causal effects on the model’s
output samples during testing, thereby evaluating the causal
reasoning of the model’s predictions. Contrary to the intuitive expectation that a correct CoT always leads to a correct answer (and an incorrect CoT always leads to an incorrect answer), our tests revealed a disproportionately high frequency of correct answers following incorrect CoTs (and vice versa). 

This phenomenon implied that the chain-based reasoning approach may have inherent attentional-defocus flaws that are difficult to resolve. Therefore, we have designed a causally enhanced method based on Treatment Effect\cite{yao2021survey} assessments to solve LLMs' reasoning problems without relying on chain-based paradigms, i.e., CSCE. This method enhances the LLM's causal reasoning by incorporating Individual Treatment Effect (ITE) metrics, including its absolute expectation and variance, into the training loss function. This integration can simultaneously improve the causal significance and consistency of the LLM's inner inference process by directly optimizing causal relationships during training\cite{ATE1, ATE2, ATE3, ITE1, ITE2, ITE3}. By explicitly modeling and optimizing for causal effects, CSCE is more focused on the underlying causal structure rather than relying on mere correlations, thereby reducing the likelihood of causal hallucinations. 

\begin{figure*}[t]
    \centering
    \vspace{-15pt}
    \includegraphics[width=1\textwidth]{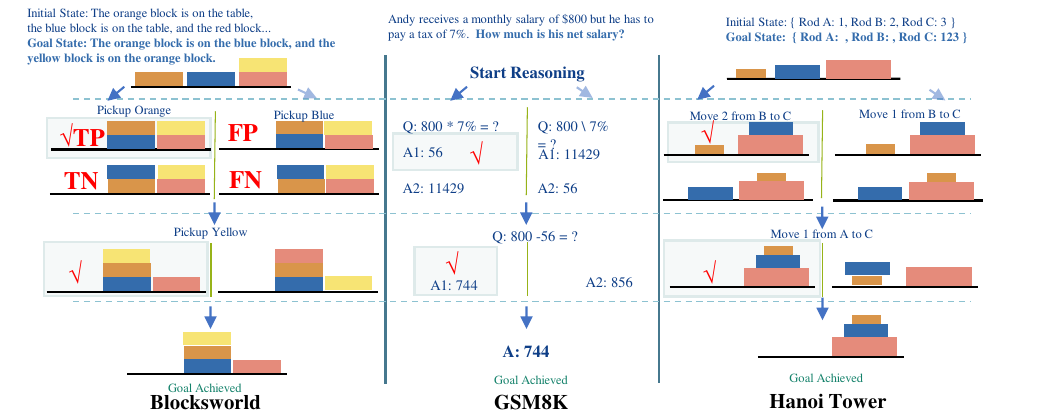}
    \vspace{-10pt}
    \caption{Task Scenario Diagram. The Blocksworld task aims to rearrange blocks from an initial state to a specified configuration through a series of actions. The GSM8K task involves mathematical reasoning, where the model computes a solution to a multi-step arithmetic problem. The Hanoi Tower task tests the model’s ability to move disks between rods while adhering to specific rules, which, based on Blocksworld, have added strict stacking order restrictions.}
    \vspace{-18pt}
    \label{fig:main}
\end{figure*}

We tested our approach against three chain-based methods—CoT\cite{CoT}, RoT\cite{RoT}, and RAP\cite{RAP}—on the Blocksworld\cite{Blocksworld}, GSM8K\cite{GSM8K}, and Hanoi Tower\cite{HanoiTower} datasets, as shown in Fig.\ref{fig:main}. 

The main contributions of this paper are as follows: 

\textbf{1. }We propose a framework for training LLMs with causal enhancements based on Treatment Effect assessments, greatly mitigating the issue of causal hallucinations featuring the weak causal relationship between a step of reasoning and corresponding state transitions. 

\textbf{2. }We effectively integrate LLMs' traditional cross-entropy loss function with the absolute expectation and variance of the ITE, resulting in outputs with higher causal significance and consistency. 

\textbf{3. }We provide a theoretical analysis of introducing Treatment Effect-related methods from the field of causal inference during the unfrozen training of large models, revealing that causal consistency is also crucial in LLM reasoning in addition to causal significance.

\textbf{4. }Experiments demonstrate that our method achieves state-of-the-art (SOTA) performance in reasoning success rate and outperforms comparison methods in reasoning speed.

\section{Related Work}

\subsection{Reasoning based on LLMs}

Current LLM reasoning approaches can be broadly categorized into frozen and unfrozen models. Frozen models refer to those where the model is not typically trained further, but instead, external methods are used to guide the model's output, such as CoT\cite{CoT}, RoT\cite{RoT}, RAP\cite{RAP}, ToT\cite{ToT}, and GoT\cite{GoT}. These methods, known as Chain-Based methods, break down the problem into a series of cascading reasoning processes to query the model. On the other hand, Unfrozen models enhance the model through fine-tuning pre-trained models rather than relying solely on external guidance, like ICL\cite{ICL}, SFT\cite{SFT}, and RLHF\cite{RLHF}. Both approaches can be used to enhance the reasoning capabilities of LLMs\cite{wang2024credes,lens.org/118-436-039-374-998}. Our method belongs to the unfrozen model category.

\subsection{Causal inference method for LLM training}
Causality is initially divided into three levels\cite{3-level}, with the Treatment Effect\cite{TreatmentEffect} widely used as an evaluation metric in machine learning. By leveraging causal analysis, a method used to identify and understand the reasons and effects of different behaviors or decisions\cite{cau-1,cau-2}, we can assist in training LLMs for reasoning tasks. This involves examining the reasons or causes behind specific events and the potential outcomes they may produce. By analyzing the differences between observational and interventional distributions, as well as evaluating the causal significance and consistency between a step of reasoning and corresponding state transitions that guide the correct answer output\cite{cau-d}, we achieve Causal Enhancement of LLM reasoning without relying on Chain-Based strategies.

\section{Method}
This section will discuss how to implement causal enhancement in the LLM reasoning process based on the Treatment Effect and the theoretical analysis behind this approach. We will also briefly compare it with chain-based methods.

\subsection{CSCE Framework Overview}

\begin{figure}[htbp]
\vspace{-10pt}
\centerline{\includegraphics[width=1\columnwidth]
{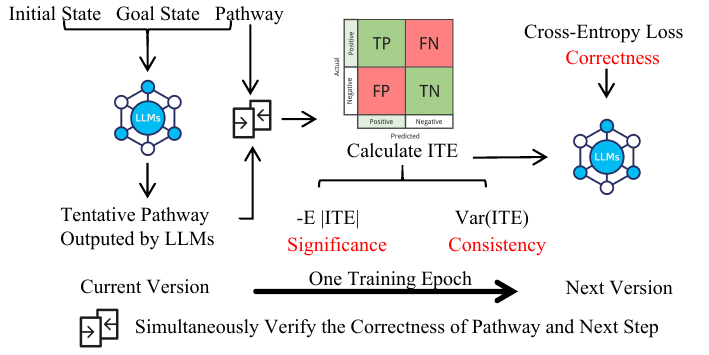}}
\caption{Schematic Diagram of CSCE Framework. In this figure, \textit{step} refers to the example steps of problems (as shown in Fig.\ref{fig:main}) that must be solved in the training set, and \textit{version} refers to every checkpoint during model training, which will generate the tentative pathway and next step. The \textit{CSCE loss} considers the cross-entropy Loss, which uses functions from the original pre-trained model, and \textit{ITE-based loss}, which is composed of two parts as shown in this figure.}
\vspace{-15pt}
\label{fig0}
\end{figure}

The framework we designed is shown in Fig.\ref{fig0}. It differs from chain-based methods like CoT, which our model generates the entire sequence of steps in one go during the reasoning validation phase without relying on externally provided prompts from a chain. Specifically, we designed a loss function that better controls the training process by evaluating the causal significance and consistency between adjacent reasoning steps and state transitions. This is a real-time adjustment process during training, and the enhanced pre-trained model will use the same loss function and parameters during the reasoning validation phase. The detailed design of the loss function will be elaborated in III.C.

\subsection{Training LLMs with ITE-Based Causal Enhancement}
In causal inference, ITE measures the difference in outcomes for an individual with and without a specific treatment. A larger ITE typically indicates a stronger causal relationship between random variables. Its definition is as follows: 
\begin{equation}
\begin{split}
\text{ITE}_i =   Y_i(W = 1) - Y_i(W = 0) 
\end{split}
\end{equation}
where \(Y_i(W = 1)\) and \(Y_i(W = 0)\) are the potential treated/control outcomes of sample \(i\). \(W\) represents the treatment assignment.

In this section, we demonstrate that merely controlling the expectation of ITE is insufficient to enhance causal significance; controlling \( \text{Var}(\mathrm{ITE}) \) is also crucial for consistency.

\begin{figure}[htbp]
\vspace{-10pt}
\centerline{\includegraphics[width=1\columnwidth]{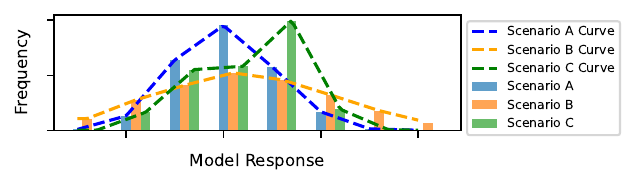}}
\vspace{-15pt}
\caption{Model Response Diagram. }
\vspace{-20pt}
\label{fig3}
\end{figure}

Based on the Fig.\ref{fig3}, we conducted a statistical analysis of the distribution of the model's output results, which demonstrates that these outputs include various possibilities, such as True Positives (TP), False Positives (FP), and False Negatives (FN), as shown in Fig.\ref{fig0}. Previous work has shown that  LLMs possess basic logical reasoning abilities, so we aim to enhance this capability rather than rebuild it. The model's responses follow an approximate normal distribution \(\text{ITE}_i \sim N(\mu, \sigma^2)\) for repeated experiments on a single sample\cite{nor1,nor2,nor3}. In this context, the mean of the normal distribution aligns with the causal significance for individual-level \(\text{ITE}_i\), while the variance reflects the causal consistency of individual effects. Based on this, we propose the following logical extension, as is shown in Fig.\ref{fig3}:

Scenario A: When individual effect consistency is high (small \(\sigma^2\)), the causal effect is more robust and consistent, resulting in a strong causal relationship even if the ITE is low. Scenario B: When individual effect consistency is low (large \(\sigma^2\)), a high ITE does not necessarily indicate a strong causal relationship because the variability among individual responses may be significant. Scenario C: When both individual effect consistency and ITE are high. In this scenario, we assert that the causal relationship is strong.

As Scenario C is what we chosen, Fig. 3 has revealed that, incorporating both the expected causal effect \( E(\text{ITE}) \) and the variance \( \text{Var}(\text{ITE}) \) during model training helps to enhance the model’s ability to capture not only the average causal effect but also the consistency across different scenarios. The expected value \( E(\text{ITE}) \) reflects the overall direction of the causal relationship, guiding the model to focus on the most significant causal changes. On the other hand, the variance \( \text{Var}(\text{ITE}) \) indicates the stability of the causal effect across different samples, promoting robustness and reducing overfitting. By optimizing both \( E(\text{ITE}) \) and \( \text{Var}(\text{ITE}) \) jointly, the model can achieve a better balance between understanding the general causal trends and ensuring consistency in predictions, thus facilitating the transition between different scenarios (e.g., from A and B to C). This approach, in combination with cross-entropy to ensure model correctness, enhances model performance\cite{louizos2017causal}.

\subsection{Loss Design}
We use $\mathrm{ITE}$ to quantitatively estimate the causal relationship between each reasoning step and the corresponding state transitions. In CSCE, we incorporate $\mathrm{ITE}$ into the training loss function alongside cross-entropy, enhancing these state transitions' causal significance and consistency. The official pre-trained large model's inherent cross-entropy loss governs the reasoning path selection, while the $ITE$ loss is focused on suppressing hallucinations. By jointly considering both causal significance $\mathrm{|E(ITE)|}$ and causal consistency \( \text{Var}(\mathrm{ITE}) \), we achieve a balance that improves model performance. Additionally, we use perplexity (PPL) as a performance metric for LLMs, with lower values indicating better predictive accuracy.

Given two binary variables \( P \) and \( Q \), which represent the correctness of each reasoning step and the corresponding state transitions respectively, where \( P, Q \sim B(0,1) \) and \( P=1 \) (or \( Q=1 \)) indicates correctness. We start by calculating the cause-effect interventions between \( P \) and \( Q \). Subsequently, we modify the distribution of \( Q \) by intervening in \( P \). From a statistical correlation perspective, if \( P \) and \( Q \) are correlated, \( Q \) can be predicted using \( P \). However, if there is no causal relationship between \( P \) and \( Q \), intervening in \( P \) will not alter the distribution of \( Q \). Therefore, even if \( P \) and \( Q \) are correlated, without a causal connection, manipulating or intervening in \( P \) will not affect the distribution of \( Q \). This distinction is essential in statistical analysis and experimental design, as it addresses the common misconception that correlation implies causality.

We aim to establish a one-to-one correspondence between a step of reasoning and corresponding state transitions. It is important to note that this one-to-one correspondence does not necessarily imply that the combination of each reasoning step and its corresponding state transition is correct. During the training process of an llm, the model's own cross-entropy loss function also plays a crucial role. Similar to methods like RAP and CoT, which freeze the model, the cross-entropy loss function can partially ensure consistency. Moreover, since our model is not frozen, its output consistency improves with each training epoch.

ITE does not operate in isolation but works in conjunction with the cross-entropy loss function. The dimensions of cross-entropy loss and ITE are different. In the early stages of model training, the changes in cross-entropy loss (or PPL) are much more significant than the changes in ITE, minimizing ITE's influence at this stage. However, in the later stages of training, as the cross-entropy loss decreases and its rate of change approaches that of ITE, ITE plays a more significant role, assisting the model in further optimization.

Consequently, we incorporate the $\mathrm{ITE}$ into the loss function, as is shown in \eqref{def}, \(p_{1|P}\) and \(p_{0|P}\) denote the conditional probabilities of \(Q\) being \emph{1} and \emph{0}, respectively, given the state of \(P\).
\begin{equation}
\mathcal{L}_{CrossEntropy}= -\left[ Q \log(p_{1|P}) + (1-Q) \log(p_{0|P}) \right] \label{abc}
\end{equation}

\begin{equation}
\mathcal{L}_{Loss} = \mathcal{L}_{CrossEntropy} -\alpha \mathrm{|E(ITE)|} + \beta \mathrm{Var(ITE)}= \mathrm{\ln(PPL)} \label{def}
\end{equation}

\section{Experiment}
\subsection{Experimental Scenarios}
In our experiments, we utilized three key datasets: Blocksworld\cite{Blocksworld}, GSM8K\cite{GSM8K}, and a custom-made Hanoi Tower dataset, which are briefly introduced in Fig.\ref{fig:main}. Blocksworld involves stacking \(n\) blocks in a specified order, with the LLM performing actions like picking up, putting down, unstacking, and stacking blocks, all while manipulating only one block at a time. GSM8K comprises 1,319 grade school math problems requiring multi-step calculations, which we address by decomposing each problem into smaller sub-questions. Our Hanoi Tower dataset, constructed with random initial and goal states, is more complex than Blocksworld. It incorporates a classical solving algorithm that translates solution paths into textual data, similar to Blocksworld, but with an added complexity of stack order judgment. Errors in the stack order result in solving failures, making the task more challenging. Unlike Blocksworld, where solution steps vary, the steps in our Hanoi Tower dataset are always odd and aligned with the minimum number of required moves. The higher complexity of the Hanoi Tower problem, due to its stricter step constraints and exponential growth of solution paths, influences the experimental setup. Despite having a similar dataset size to Blocksworld, the increased difficulty of Hanoi Tower leads to more challenging evaluation, impacting the model's performance and the generalization of results.
\subsection{Baseline}
In our study, we utilized several pre-trained models as baselines, including LLAMA-2-7B\cite{Llama-2}, Phi-2-7B\cite{phi-2}, Mistral-7B\cite{Mistral} and Mixtral-8x7B\cite{Mixtral}. The training primarily involved 7B models on a single NVIDIA A100 GPU, with models loaded in 4-bit. We also evaluated three techniques to enhance reasoning capabilities: RAP, which employs Monte Carlo Tree Search (MCTS) for strategic exploration\cite{RAP}, transforming LLMs into reasoning agents and world models; CoT\cite{CoT}, which improves reasoning by generating intermediate reasoning steps; and RoT\cite{RoT}, a framework enhancing tree-search-based prompting methods, leveraging past experiences to improve reasoning and planning.

\subsection{Hyperparameters}
In our study, all training processes were conducted using the original hyperparameters provided by the official releases of the respective pretrained models, without any modifications. Specifically, we followed the default configurations for each model during training to ensure consistency with the original models. These hyperparameters include key settings such as learning rate, batch size, optimizer, and number of training epochs, all of which were kept in line with the values recommended in the official documentation.
\subsection{Results}
We acknowledge that our experiments were primarily conducted on 7B parameter models due to computational constraints and availability. Despite this, our method achieved high accuracy rates across the datasets categorized as \textit{Structured Problems}, including Blocksworld and Hanoi Tower, consistently outperforming baseline methods. The use of a 7B model is sufficient to enhance the effectiveness of our approach, as demonstrated by its superior success rates in these tasks, as shown in Table~\ref{table1}. In the Blocksworld tasks, our method exhibited strong inference capabilities. In contrast, our model maintained higher accuracy than other models on the more complex Hanoi Tower tasks, even under stricter stacking order requirements. This robustness across different structured problem types underscores the reliability of our approach.

\begin{table}[htbp]
\vspace{-10pt}
\caption{Success Rate Under Structured Problems}
\vspace{-15pt}
\centering
\begin{center}
\resizebox{\columnwidth}{!}{%
\begin{tabular}{|l|c|c|c|c|c|c|}
\hline
\textbf{Task} & \multicolumn{3}{c|}{\textbf{Blocksworld}} & \multicolumn{3}{c|}{\textbf{Hanoi Tower}} \\
\cline{2-7}
\textbf{Model \& Method} & \textbf{\textit{2-Step}} & \textbf{\textit{4-Step}} & \textbf{\textit{6-Step}} & \textbf{\textit{3-Step}} & \textbf{\textit{5-Step}} & \textbf{\textit{7-Step}} \\
\hline
Llama-2-7B + RAP  & 0.39 & 0.41 & 0.37 & 0.29 & 0.21 & 0.11 \\
Llama-2-7B + CoT & 0.50 & 0.63 & 0.40 & 0.34 & 0.23 & 0.10 \\
Llama-2-7B + RoT & 0.52 & 0.67 & 0.27 & 0.41 & 0.27 & 0.13 \\
Llama-2-7B + \textbf{Ours} & \textbf{0.95} & \textbf{0.80} & \textbf{0.76} & \textbf{0.45} & \textbf{0.39} & \textbf{0.24} \\
\hline
Phi-2-7B + RAP & 0.40 & 0.44 & 0.33 & 0.27 & 0.21 & 0.14 \\
Phi-2-7B + CoT & 0.43 & 0.05 & 0.01 & 0.33 & 0.22 & 0.10 \\
Phi-2-7B + RoT & 0.54 & 0.16 & 0.01 & 0.24 & 0.12 & 0.02 \\
Phi-2-7B + \textbf{Ours} & \textbf{0.91} & \textbf{0.86} & \textbf{0.79} & \textbf{0.40} & \textbf{0.25} & \textbf{0.17} \\
\hline
Mistral-7B + RAP & 0.49 & 0.41 & 0.35 & 0.34 & 0.25 & 0.14 \\
Mistral-7B + CoT & 0.84 & 0.41 & 0.24 & 0.40 & 0.32 & 0.21 \\
Mistral-7B + RoT & 0.81 & 0.49 & 0.21 & 0.35 & 0.22 & 0.17 \\
Mistral-7B + \textbf{Ours} & \textbf{0.97} & \textbf{0.94} & \textbf{0.82} & \textbf{0.49} & \textbf{0.37} & \textbf{0.26} \\
\hline
Mixtral-8x7B + RAP & 0.49 & 0.44 & 0.35 & 0.40 & 0.24 & 0.15 \\
Mixtral-8x7B + CoT & 0.81 & 0.63 & 0.55 & 0.45 & 0.27 & 0.14 \\
Mixtral-8x7B + RoT & 0.87 & 0.71 & 0.55 & 0.37 & 0.22 & 0.10 \\
Mixtral-8x7B + \textbf{Ours} & \textbf{0.99} & \textbf{0.97} & \textbf{0.93} & \textbf{0.50} & \textbf{0.35} & \textbf{0.22} \\
\hline
\end{tabular}%
}
\vspace{-10pt}
\label{table1}
\end{center}
\end{table}

Additionally, we verified our method's capabilities on the GSM8K dataset, where it outperformed baseline methods like RAP, RoT, and CoT. This further confirms our method's advantage in completing multi-step reasoning tasks, demonstrating its ability to excel in solving highly structured problems and enhancing LLMs' performance in unstructured mathematical problem-solving, as is shown in Table~\ref{table2}. The Mixtral-8x7B model, being larger, serves as a reference for understanding the potential performance at larger scales.
\begin{table}[htbp]
\vspace{-10pt}
\caption{Success Rate Under Unstructured Problems}
\vspace{-10pt}
\begin{center}

\begin{tabular}{|l|c|c|c|c|}
\hline
\textbf{Task} & \multicolumn{4}{c|}{\textbf{GSM8K}} \\
\cline{2-5}
\textbf{Model \& Method} & \textbf{\textit{RAP}} & \textbf{\textit{RoT}} & \textbf{\textit{CoT}} & \textbf{\textit{Ours}} \\
\hline
Llama-2-7B  & 0.51 & 0.54 & 0.47 & \textbf{0.92} \\
Phi-2-7B    & 0.45 & 0.48 & 0.48 & \textbf{0.89} \\
Mistral-7B  & 0.39 & 0.32 & 0.31 & \textbf{0.85} \\
Mixtral-8x7B & 0.48 & 0.50 & 0.49 & \textbf{0.90} \\
\hline
\end{tabular}
\vspace{-10pt}
\label{table2}
\end{center}
\end{table}

To verify the success rate of our CSCE method on other baseline tasks, we designed a control experiment to ensure that our approach does not impair the model's inherent problem-solving and reasoning abilities. The experimental results indicate that the CSCE method can, to some extent, enhance the model's problem-solving capabilities on other baseline tasks without causing any reduction in performance. See Table~\ref{Our-Results-other-Table}.

\begin{table}[htbp]
\vspace{-10pt}
\caption{Results of Model’s Inherent Capabilities}
\begin{center}
\vspace{-5pt}
\begin{tabular}{|l|c|c|}
\hline
\textbf{Model \& Method} & \textbf{AQUA} & \textbf{QASC} \\
\hline
Llama-2-7B  & 0.25  & 0.17 \\
Llama-2-7B + CSCE & \textbf{0.74}  & \textbf{0.62} \\
\hline
Baichuan-7B  & 0.31  & 0.07 \\
Baichuan-7B + CSCE & \textbf{0.85}  & \textbf{0.31} \\
\hline
Mpt-7B   & 0.11  & 0.05 \\
Mpt-7B + CSCE  & \textbf{0.65}  & \textbf{0.27} \\
\hline
TAIDE-LX-7B   & 0.27  & 0.21 \\
TAIDE-LX-7B + CSCE & \textbf{0.89}  & \textbf{0.72} \\
\hline
Qwen1.5-7B   & 0.57  & 0.09 \\
Qwen1.5-7B + CSCE & \textbf{0.75}  & \textbf{0.37} \\
\hline
\end{tabular}
\vspace{-10pt}
\label{Our-Results-other-Table}
\end{center}
\end{table}

\begin{figure}[htbp]
\centerline{\includegraphics[width=\columnwidth]{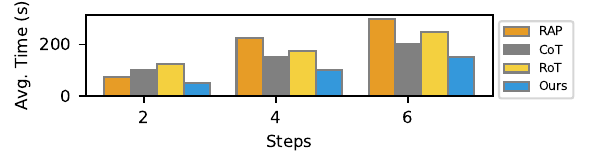}}
\vspace{-15pt}
\caption{Improvement in Reasoning Speed.}
\vspace{-15pt}
\label{fig2}
\end{figure}

Our approach significantly shortens the time required to complete long-range reasoning tasks compared to benchmarks like CoT, RoT, and RAP, as shown in Fig.\ref{fig2}. The main reason for our higher inference speed is that, unlike other methods, our approach performs simultaneous multi-step reasoning and outputs the answer in one go. In contrast, RAP relies on Monte Carlo search tree inference, where each node in the tree requires a separate inference step, leading to slower overall performance. CoT and RoT models, which rely on chained stacked inference, require multiple short inductions for a single long process, further slowing down the inference process. Our method eliminates the need for multiple inferences, making it more efficient, especially in longer reasoning scenarios.
\subsection{More Details}
In our experiments, we fine-tuned pretrained models, which is a standard practice in deep learning for adapting models to specific tasks. Fine-tuning allows the model to leverage the knowledge from the pretrained weights while adjusting to the nuances of the target task. In contrast, a frozen model, which keeps the pretrained weights unchanged, may not perform as well on specialized tasks due to its inability to adapt to the task-specific patterns and complexities.

To further evaluate the impact of our approach, we conducted ablation experiments where we removed the ITE-based modifications in the loss function. Due to the time-intensive nature of these experiments, we completed a subset of the tests. Specifically, using the Llama-2-7B model on the Blocksworld dataset, we observed success rates of 83\%, 74\%, and 44\% for tasks requiring 2, 3, and 6 steps, respectively. As expected, performance declined significantly for longer tasks, which highlights the effectiveness of the ITE-based modifications.

These findings underscore the importance of fine-tuning for task-specific improvements and demonstrate that the performance improvements we observed are attributable to the ITE-based method, rather than being a mere result of fine-tuning. Thus, while frozen models could serve as a baseline, fine-tuning with the proposed method is essential for achieving the best performance on complex tasks.
\section{Conclusion}
This study introduced a novel approach to enhance LLMs in handling structured and unstructured reasoning tasks, demonstrating significant improvements in accuracy and efficiency across various problem types. Our method proved effective in structured problem scenarios like Blocksworld and Hanoi Tower, addressing challenges such as causal hallucinations and large search spaces. However, in scenarios with strict order requirements, like Hanoi Tower, the accuracy could have been higher. It also showed strong performance in non-structured mathematical tasks, such as those found in the GSM8K dataset, further highlighting its versatility. Future work will refine the method to enhance scalability and efficiency across complex problem-solving tasks, including structured and unstructured scenarios.

\section{Appendix}

\subsection{A Note on the Hanoi Tower Dataset}
We generated and produced the Hanoi Tower dataset in the paper. The production method is to randomly generate several states conforming to the placement rules of the Hanoi Tower based on a given number of rods and disks, e.g., three rods and three disks, and randomly select one of these states as the starting and target states for a single sample. For a single sample, the classical partition algorithm is used to derive the pathway, and according to the length of the pathway, the sample is categorized into different number of steps groups, e.g., 3-steps, 5-steps, 7-steps, and so on. An odd number is chosen for the allocation because the most complex solving step of Hanoi Tower in the case of three rods and $n$ disks is $2^n-1$ steps. We generated the dataset Hanoi Tower using exactly the same storage format and Prompt structure as Blocksworld and GSM8K.

\subsection{Hyperparameters}
We all use the official hyperparameters of the pre-trained model, and hyperparameter tuning is not addressed in this paper.

\subsection{Prompt Templates Used During Training and Testing of CSCE}

\floatname{algorithm}{Prompt}
\setcounter{algorithm}{0}
\setlength{\textfloatsep}{-20pt}  

\begin{algorithm}[H]  
\caption{Prompt Templates Used During \textbf{Training}}

\begin{algorithmic}
\State \textbf{Input:} \textbf{Initial State} \texttt{||}  \textbf{ Goal State} \#\#\#\# \textbf{ Pathway}
\State \textbf{Output:} \#\#\#\# Pathway
\State \textbf{Pathway:} \texttt{<Step1><Step2><Step3><Step4>}
\end{algorithmic}

\end{algorithm}

\vspace{-15pt}  

\begin{algorithm}[H]  
\caption{Prompt Templates Used During \textbf{Testing}}

\begin{algorithmic}
\State \textbf{Input:} \textbf{Initial State} \texttt{||} \textbf{ Goal State}
\State \textbf{Output:} \#\#\#\# Pathway
\State \textbf{Pathway:} \texttt{<Step1><Step2><Step3><Step4>}
\end{algorithmic}

\end{algorithm}

\bibliographystyle{IEEEtran}
\bibliography{IEEEabrv,camera}

\vspace{12pt}

\end{document}